\documentclass{article} 
\usepackage[dvipsnames]{xcolor}
\usepackage{sigbovik2019_conference,times}
\usepackage[hidelinks]{hyperref}
\usepackage{url}
\usepackage{soul}
\usepackage{dirtytalk}
\usepackage{epigraph}
\usepackage{subfig}
\usepackage{listings}
\usepackage{amsmath}
\usepackage{graphicx}
 
\definecolor{codegreen}{rgb}{0,0.6,0}
\definecolor{codegray}{rgb}{0.5,0.5,0.5}
\definecolor{codepurple}{rgb}{0.58,0,0.82}
\definecolor{backcolour}{rgb}{0.95,0.95,0.92}

\lstdefinestyle{pystyle}{
    backgroundcolor=\color{backcolour},   
    commentstyle=\color{codegray},
    keywordstyle=\color{codegreen},
    numberstyle=\tiny\color{codegray},
    stringstyle=\color{codepurple},
    basicstyle=\ttfamily\footnotesize,
    breakatwhitespace=false,         
    breaklines=true,                 
    captionpos=b,                    
    keepspaces=true,                 
    numbers=left,                    
    numbersep=5pt,                  
    showspaces=false,                
    showstringspaces=false,
    showtabs=false,                  
    tabsize=2
}

\title{Deep Industrial Espionage}

\author{Samuel Albanie, James Thewlis, Sebastien Ehrhardt \& Jo\~{a}o F. Henriques \\
Dept. of Deep Desperation,\\
UK (EU at the time of submission) \\
}

\iclrfinalcopy 

\begin{document}
\maketitle


\begin{abstract}
The theory of deep learning is now considered largely solved, and is well understood by researchers and influencers alike.  To maintain our relevance, we therefore seek to apply our skills  to under-explored, lucrative applications of this technology. To this end, we propose and \textit{Deep Industrial Espionage}, an efficient end-to-end framework for industrial information propagation and productisation.  Specifically, given a single image of a product or service, we aim to reverse-engineer, rebrand and distribute a copycat of the product at a profitable price-point to consumers in an emerging market---all within in a single forward pass of a Neural Network.    Differently from prior work in machine perception which has been restricted to classifying, detecting and reasoning about object instances, our method offers \textit{tangible business value} in a wide range of corporate settings.  Our approach draws heavily on a promising recent arxiv paper until its original authors' names can no longer be read (we use felt tip pen).  We then rephrase the anonymised paper, add the word \say{novel} to the title, and submit it a prestigious, closed-access espionage journal who assure us that \textit{someday}, we will be entitled to some fraction of their extortionate readership fees. 

\end{abstract}

\section{Introduction}

In the early 18th Century, French Jesuit priest François Xavier d'Entrecolles radically reshaped the geographical distribution of manufacturing knowledge.  Exploiting his diplomatic charm and privileged status, he gained access to the intricate processes used for porcelain manufacture in the Chinese city of Jingdezhen, sending these findings back to Europe (over the course of several decades) in response to its insatiable demand for porcelain dishes \citep{giaimo}.  This anecdote is typical of corporate information theft: it is an arduous process that requires social engineering and expert knowledge, limiting its applicability to a privileged minority of well-educated scoundrels.

Towards reducing this exclusivity, the objective of this paper is to democratize industrial espionage by proposing a practical, fully-automated approach to the theft of ideas, products and services.   Our method builds on a rich history of \textit{analysis by synthesis} research that seeks to determine the physical process responsible for generating an image.  However, in contrast to prior work that sought only to determine the parameters of such a process, we propose to instantiate them with a \textit{just-in-time}, minimally tax-compliant manufacturing process.  Our work points the way to a career rebirth for those like-minded members of the research community seeking to maintain their raison d'\^etre in the wake of recent fully convolutional progress. 

Concretely, we make the following four contributions: (1) We propose and develop \textit{Deep Industrial Espionage} (henceforth referred to by its cognomen, \textit{Espionage}) an end-to-end framework which enables industrial information propagation and hence advances the \textit{Convolutional Industrial Complex}; (2) We introduce an efficient implementation of this framework through a novel application of differentiable manufacturing and sunshine computing; (3)  We attain qualitatively state-of-the-art product designs from several standard corporations; (4) We sidestep ethical concerns by failing to contextualise the ramifications of automatic espionage for job losses in the criminal corporate underworld. 
\section{Related Work}

\textbf{Industrial Espionage} has received a great deal of attention in the literature, stretching back to the seminal work of \cite{prometheus} who set the research world alight with a well-executed workshop raid, a carefully prepared fennel stalk and a passion for open source manuals.   A comprehensive botanical subterfuge framework was later developed by \cite{fortune1847three} and applied to the appropriation of Chinese \textit{camellia sinensis} production techniques, an elaborate pilfering orchestrated to sate the mathematically unquenchable British thirst for tea. More recent work has explored the corporate theft of internet-based prediction API model parameters, thereby facilitating a smorgasbord of machine learning shenanigans \citep{tramer2016stealing}.  In contrast to their method, our \textit{Espionage} reaches beyond web APIs and out into the bricks and mortar of the physical business world.  Astute readers may note that in a head-to-head showdown of the two approaches, their model could nevertheless still steal our model's parameters. Touch\'{e}. 
Finally, we note that while we are not the first to propose a convolutional approach to theft \citep{boredYL}, we are likely neither the last, adding further justification to our approach.

\textbf{Analysis by Synthesis.} Much of the existing work on analysis by synthesis in the field of computer vision draws inspiration from Pattern Theory, first described by \cite{grenander}.  The jaw dropping work of \cite{blanz1999morphable} enabled a range of new facial expressions for Forrest Gump.  This conceptual approach was generalised to the OpenDR framework through the considerable technical prowess of \cite{loper2014opendr}, who sought to achieve generic end-to-end (E2E) differentiable rendering. Differently from OpenDR, our approach is not just E2E, but also B2B (business-to-business) and B2C (business-to-consumer).
\section{Method}

\begin{figure}[t]
    \vspace{-2cm}
    \hspace{-2.3cm}\includegraphics[width=1.6\textwidth]{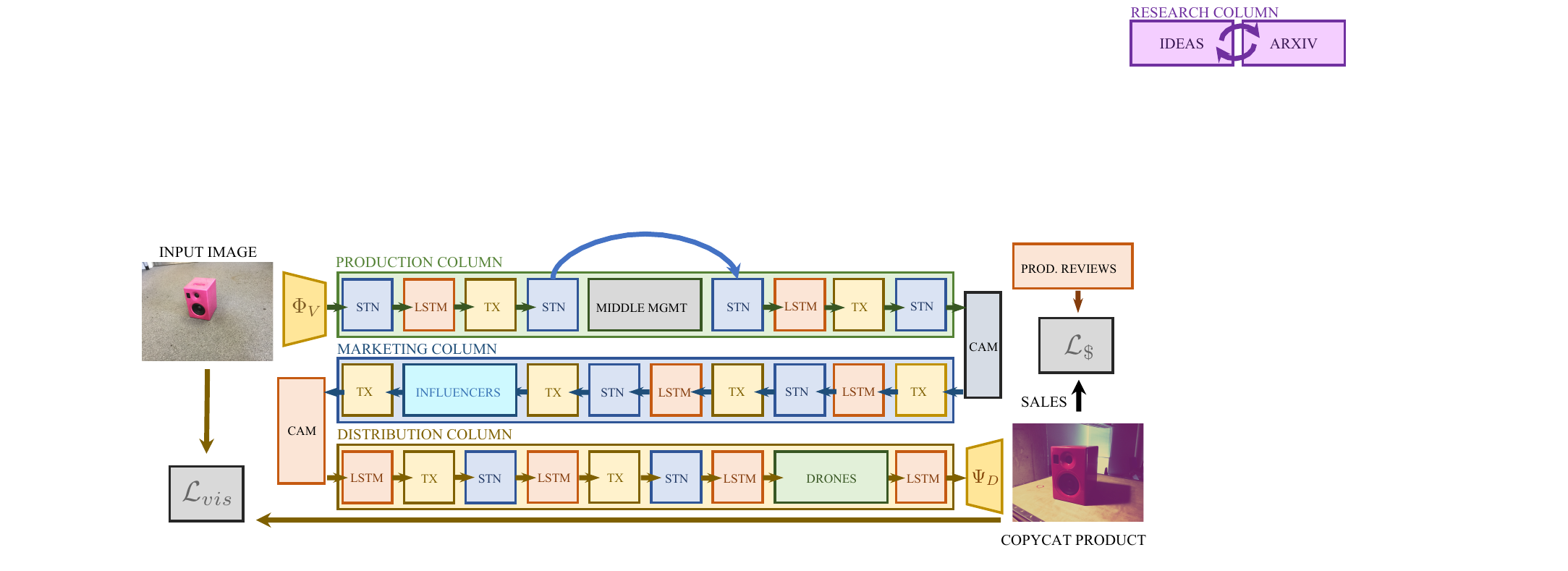}
    \caption{A random projection of the proposed multi-dimensional \textit{Espionage} architecture.  We follow best-practice and organise business units as tranposed horizontally integrated functional columns. The trunk of each column comprises stacks of powerful acronyms, which are applied following a Greek visual feature extractor $\Phi_V$. 
    Gradients with respect to the loss terms $L_{\$}$ and $L_{vis}$ flow liberally across the dimensions (see Sec. \ref{sec:diff-man} for details). We adopt a snake-like architecture, reducing the need for a rigid backbone and producing an altogether more sinister appearance. }
    \label{fig:framework}
\end{figure}

The \textit{Espionage} framework is built atop a new industrial paradigm, namely \textit{differentiable manufacturing}, which is described in Sec.~\ref{sec:diff-man}.  While theoretically and phonaesthetically pleasing, this approach requires considerable computational resources to achieve viability and would remain intractable with our current cohort of trusty laptops (acquired circa 2014).  We therefore also introduce an efficient implementation of our approach in Sec.~\ref{sec:inclement-comp} using a technique that was tangentially inspired by a recent episode of the Microsoft CMT submission gameshow while it was raining.

\subsection{Differentiable manufacturing} \label{sec:diff-man}

Recent developments in deep learning have applied the "differentiate everything" dogma to everything, from functions that are not strictly differentiable at every point (ReLU), to discrete random sampling \citep{maddison2016concrete,jang} and the sensory differences between dreams and reality. 
Inspired by the beautiful diagrams of \cite{maclaurin2015gradient}, we intend to take this idea to the extreme and perform end-to-end back-propagation through the full design and production pipeline.
This will require computing gradients through entire factories and supply chains.
Gradients are passed through factory workers by assessing them locally, projecting this assessment by the downstream gradient, and then applying the chain rule.
The chain rule only requires run-of-the-mill chains, purchased at any hardware store (fluffy pink chaincuffs may also do in a pinch), and greatly improves the productivity of any assembly line.  Note that our method is considerably smoother than \textit{continuous manufacturing}---a technique that has been known to the machine learning community since the production of pig iron moved to long-running blast furnaces.

Two dimensions of the proposed \textit{Espionage} framework is depicted in Fig.~\ref{fig:framework}.  At the heart of the system is a pair of losses, one visual, $L_{vis}$, one financial $L_{\$}$.  For a given input image, the visual loss encourages our adequately compensated supply line to produce products that bear more than a striking resemblance to the input.  This is coupled with a second loss that responds to consumer demand for the newly generated market offering.  
Our system is deeply rooted in computer vision:  thus, while the use of Jacobians throughout the organisation ensures that the full manufacturing process is highly sensitive to customer needs, the framework coordinates remain barycentric rather than customer-centric. To maintain our scant advantage over competing espionage products, details of the remaining $n - 2$ dimensions of the diagram are omitted. 

\begin{figure}
\begin{lstlisting}[language=Python]
try:  # often fails on the first import - never understood why
    from espionage import net
except Exception:  # NOQA 
    pass  # inspecting the exception will bring you no joy
if "x" in locals(): del x # DO NOT REMOVE THIS LINE
from espionage import net # if second fail, try re-deleting symlinks?
net.steal(inputs) # when slow, ask Seb to stop thrashing the NFS (again)
\end{lstlisting}
\caption{A concise implementation of our method can be achieved in only seven lines of code}
\label{fig:code}
\end{figure}

\subsection{Sun Macrosystems} \label{sec:inclement-comp}

\epigraph{Ah! from the soul itself must issue forth \\
A light, a glory, a fair luminous cloud \\
          Enveloping the Earth}{Jeff Bezos}

Differentiable manufacturing makes heavy use of gradients, which poses the immediate risk of steep costs.  The issue is exacerbated by the rise of costly cloud\footnote{Not to be confused with Claude by our French-speaking readers, according to Sebastien's account of a recent McD'oh moment.} services, which have supported an entire generation of vacuous investments, vapour-ware and hot gas.  Despite giving birth to the industrial revolution, smog and its abundance of cloud resources (see Fig.~\ref{fig:cloud} in Appendix~\ref{app:cloud}, or any British news channel), the United Kingdom, has somehow failed to achieve market leadership in this space.

Emboldened with a \say{move fast and break the Internet} attitude~\citep{fouhey2012kardashian}, we believe that it is time to reverse this trend.
Multiple studies have revealed that sunshine improves mood, disposition, and tolerance to over-sugared \emph{caipirinhas}.
It is also exceedingly environmentally friendly, if we ignore a few global warming hiccups.\footnote{Up to about 5 billion AD, when the Sun reaches its red giant phase and engulfs the Earth.} The question remains, how does this bright insight support our grand computational framework for \textit{Espionage}?  To proceed, we must first consider prior work in this domain.

An early example of sunshine computing is the humble sundial. This technology tells the time with unrivalled accuracy and reliability, and automatically implements "daylight saving hours" with no human intervention.
Sunshine-powered sundials are in fact part of a new proposal to replace atomic clocks in GPS satellites (patent pending).
With some obvious tweaks, these devices can form the basis for an entire sunshine-based ID-IoT product line, with fast-as-light connectivity based on responsibly-sourced, outdoors-bred photons.
This is not to be confused with the electron-based "fast-as-lightning" transmission of cloud computing, an expression coined by the cloud computing lobbyists in a feeble attempt to suggest speed.  

The cloud lobby has been raining on our parade for too long and it is time to make the transition. We proceed with no concrete engineering calculations as to whether this is viable, but instead adopt a sense of sunny optimism that everything will work out fine.  Thus, with a blue sky above, sandals on our feet and joy in hearts, we propose to adopt a fully solar approach to gradient computation.

\subsection{Implementation through a Unicorn Startup}

The appearance of rainbows through the interaction of legacy cloud computing and novel sunshine computing suggests that our framework can easily attain unicorn status.
Because branding is everything, our first and only point of order was to choose the aforementioned rainbow as our logo and basis for marketing material.
This cherished symbol expresses the diversity of colours that can be found in hard cash\footnote{For the most vibrant rainbow we conduct all transactions in a combination of Swiss Francs and Australian Dollars}.

A quick back-of-the-envelope calculation showed that our startup's VC dimension is about 39 Apples, shattering several points, hopes and dreams. This quantity was rigorously verified using the advanced accounting analytics of a 40-years-old, 100MB Microsoft Excel spreadsheet that achieved semi-sentience in the process.

\begin{figure}
\centering
\begin{tabular}{llllll} 
\includegraphics[width=0.16\textwidth]{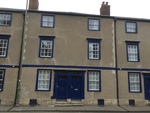}%
\includegraphics[width=0.16\textwidth]{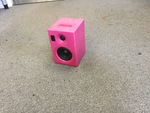}%
\includegraphics[width=0.16\textwidth]{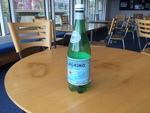}%
\includegraphics[width=0.16\textwidth]{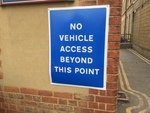}%
\includegraphics[width=0.16\textwidth]{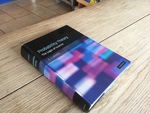}%
\includegraphics[width=0.16\textwidth]{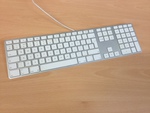}\\
\includegraphics[width=0.16\textwidth]{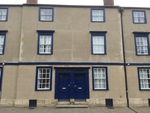}%
\includegraphics[width=0.16\textwidth]{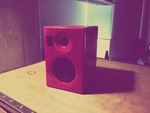}%
\includegraphics[width=0.16\textwidth]{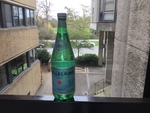}%
\includegraphics[width=0.16\textwidth]{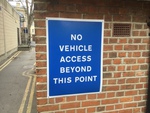}%
\includegraphics[width=0.16\textwidth]{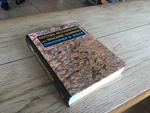}%
\includegraphics[width=0.16\textwidth]{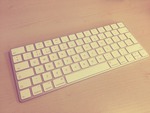}%
\end{tabular}
\caption{\textbf{Top row}: A collection of unconstrained, natural images of products. \textbf{Bottom row}: Photographs of the physical reconstructions generated by our method.  Note that the proposed \textit{Espionage} system can readily produce full houses, speakers, water bottles and street signs---all from a single image sample.  When generating books, \textit{Espionage} does not achieve an exact reconstruction, but still seeks to preserve the philosophical bent.  \textbf{Failure case}: the precise layout of keys in technology products such as keyboards are sometimes altered.}
\label{fig:exp}
\end{figure}

\section{Experiments}
  
Contemporary researchers often resort to the use of automatic differentiation in order to skip writing the backward pass, in a shameful effort to avoid undue mathematical activity. We instead opt to explicitly write the backward pass and employ symbolic integration to derive the forward pass. Thanks to advances in computational algebra~\citep{wolfram2013computer}, this method almost never forgets the $+C$.  Our method can then be implemented in just seven lines of Python code (see Fig. \ref{fig:code}).

To rigorously demonstrate the scientific contribution of our work, we conducted a large-scale experiment on a home-spun dataset of both branded and unbranded products. Example outcomes of this experiment can be seen in Fig.~\ref{fig:exp}. 

Efficacy was assessed quantitatively through a human preference study.  Unfortunately, lacking both US and non-US credit cards,  we were unable to procure the large sample pool of Amazon Mechanical Turks required to achieve statistically significant results.  We therefore turned to our immediate family members to perform the assessments.  To maintain the validity of the results, these experiments were performed doubly-blindfolded, following the rules of the popular party game ``pin the tail on the donkey''.  The instructions to each blood relative stated simply that if they loved us, they would rate the second product more highly than the first.  While there was considerable variance in the results, the experiment was a conclusive one, ultimately demonstrating both the potential of our approach and the warm affection of our loved-ones.  Comparisons to competing methods were conducted, but removed from the paper when they diminished the attractiveness of our results. 

\textbf{Reproducibility:} Much has been written of late about the nuanced ethics of sharing of pretrained models and code by the sages of the field (see e.g. \cite{openAI} and \cite{zach} for complementary perspectives).  As adequately demonstrated by the title of this work, we are ill-qualified to contribute to this discussion, choosing instead to fall back to the tried and true research code release with missing dependencies, incorrectly set hyper-parameters, and reliance on the precise ordering of \texttt{ls} with Linux Kernel 2.6.32 and ZFS v0.7.0-rc4.  This should allow us replace public concern about our motives with pity for our technical incompetence.
\section{Conclusion}

The theory of deep learning may be solved but the music need not stop.  In this work, we have made a brief but exciting foray into a new avenue of career opportunities for deep learning researchers and enthusiasts.  Nevertheless, we acknowledge that there may not be room enough for us all in the espionage racket and so we also advocate responsible preparation for the bitter and frosty depths of the upcoming AI employment winter.  To this end, we have prepared a new line of reasonably priced researcher survival kits---each will include a 25-year supply of canned rice cakes, a handful of pistachios, a ``best hit'' compilation of ML tweets in calendar form, and an original tensor-boardgame, a strategic quest for the lowest loss through the trading of GPUs and postdocs. Collectively, these items will keep spirits high and bring back fond memories of those halycon days when all that was required to get a free mug was a copy of your r\'esum\'e.  The kits will be available to purchase shortly from the dimly lit end of the corporate stands at several upcoming conferences.  


\bibliographystyle{iclr2019_conference}
\bibliography{iclr2016_conference}

\appendix

\pagebreak
\section{Appendix} \label{app:cloud}

\begin{figure}[h]
    \centering
    \includegraphics[width=0.8\textwidth]{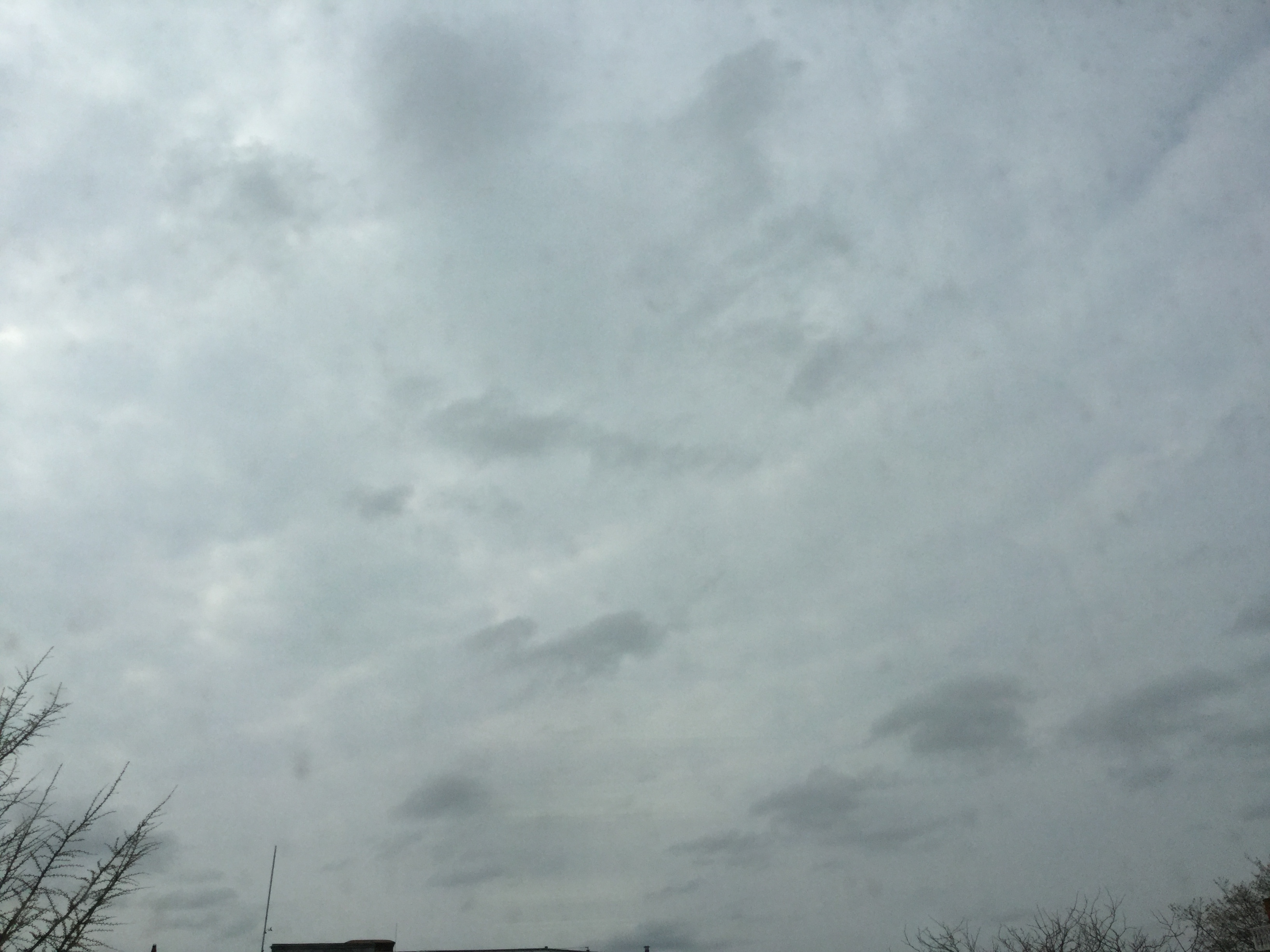}
    \caption{UK cloud-cover at the time of submission.}
    \label{fig:cloud}
\end{figure}

\end{document}